\documentclass[10pt,twocolumn,letterpaper]{article}

\usepackage[pagenumbers]{cvpr} 
\usepackage{caption}
\usepackage{graphicx}
\usepackage{amsmath}
\usepackage{amssymb}
\usepackage{booktabs}
\usepackage{xspace}
\usepackage{comment}

\usepackage[pagebackref,breaklinks,colorlinks]{hyperref}
\usepackage[dvipsnames]{xcolor}  
\usepackage[accsupp]{axessibility} 

\usepackage[capitalize]{cleveref}
\crefname{section}{Sec.}{Secs.}
\Crefname{section}{Section}{Sections}
\Crefname{table}{Table}{Tables}
\crefname{table}{Tab.}{Tabs.}

\newcommand{\td}[1]{{\textcolor{black}{#1\xspace}}}
\newcommand{\edit}[1]{{\textcolor{black}{#1\xspace}}}

\newcommand{\howtosign}{How2Sign\xspace}
\newcommand{\bslcorpus}{\textsc{BslCorpus}\xspace}
\newcommand{\phoenixT}{PHOENIX2014T\xspace}

\newcommand{\methodName}{\textsc{Spot-Align}\xspace}

\newcommand{\videoGallery}{\mathcal{V}}
\newcommand{\textGallery}{\mathcal{T}}
\newcommand{\signEncoder}{\phi_{\textsc{v}}}
\newcommand{\textEncoder}{\phi_{\textsc{t}}}
\newcommand{\textToSign}{\textsc{t2v}\xspace}
\newcommand{\signToText}{\textsc{v2t}\xspace}
\newcommand{\signVideoEmbedding}{\psi_v}
\newcommand{\queryType}{free-form\xspace}

\def\sepappendix{0}

\begin{document}

\title{Sign Language Video Retrieval with Free-Form Textual Queries}

\author{Amanda Duarte$^{1,2}$
\and
Samuel Albanie$^3$
\and
Xavier Gir{\'o}-i-Nieto$^{1,4}$
\and
G{\"u}l Varol$^{5}$ \\ \vspace{-0.4cm}
\and
$^1$\emph{Universitat Polit\`{e}cnica de Catalunya, Spain} \quad $^2$\emph{Barcelona Supercomputing Center, Spain} \quad \\
$^3$\emph{Department of Engineering, University of Cambridge, UK}\\
$^4$\emph{Institut de Rob\`{o}tica i Inform\`{a}tica Industrial, CSIC-UPC, Spain} \\
\quad $^5$\emph{LIGM, {\'E}cole des Ponts, Univ Gustave Eiffel, CNRS, France}\\
\small{\url{https://imatge-upc.github.io/sl_retrieval/}}
}

\maketitle

\begin{abstract}
Systems that can efficiently search collections of sign language videos
have been highlighted as a useful application of sign language technology.
However, the problem of searching videos beyond individual keywords 
has received limited attention in the literature.
To address this gap, in this work we introduce the task of sign language retrieval with \queryType
\footnote{
The terminology ``natural language query'' is commonly
used to describe unconstrained textual queries in spoken
languages.
However, since sign languages are also natural
languages, we adopt for the term
``free-form textual query'' instead. }
textual queries:
given a written query (e.g.\ a sentence)
and a large collection of sign language videos, the objective is to find the signing video 
that best matches the written query.
We propose to tackle this task by learning
cross-modal embeddings on the recently introduced large-scale
\howtosign dataset of American Sign Language (ASL).
We identify that a key bottleneck in the performance
of the system is the quality of the sign video embedding
which suffers from a scarcity of labelled training data.
We, therefore, propose \methodName, a framework for interleaving iterative rounds of
sign spotting and feature alignment to expand the scope and scale of available training data.
We validate the effectiveness of \methodName for learning a robust
sign video embedding through improvements in both sign recognition
and the proposed video retrieval task.

\end{abstract}
\section{Introduction}
\label{sec:intro}

Sign languages are the primary means of communication among deaf
communities.
They are visual, complex, evolved languages that employ combinations of 
\emph{manual} and \emph{non-manual markers} such as movements
of the face, body and hands to convey information~\cite{ASLstructure}.

Recent developments in automatic speech recognition (ASR) 
for spoken languages~\cite{chorowski2015attention,chan2016listen,zhang2020pushing,xu2021self} 
have enabled automatic captioning of vast swathes of
video content hosted on platforms such as YouTube. 
In addition to rendering the videos more accessible,
this captioning yields a second important benefit:
it allows the content of the videos to be indexed
and efficiently searched with text queries.
By contrast,
the same automatic captioning capability
(and hence searchability)
does not
exist for sign language content.
Indeed, recent work has drawn attention to the pressing need
to develop systems that can index archives of sign language
videos to render them searchable~\cite{bragg2019sign}.
Without these tools, sign language video creators must type the spoken language translation of their content if they want to reach the same discoverability as their spoken language counterparts.

\begin{figure}
    \centering
    \resizebox{\linewidth}{!}{
    \includegraphics[]{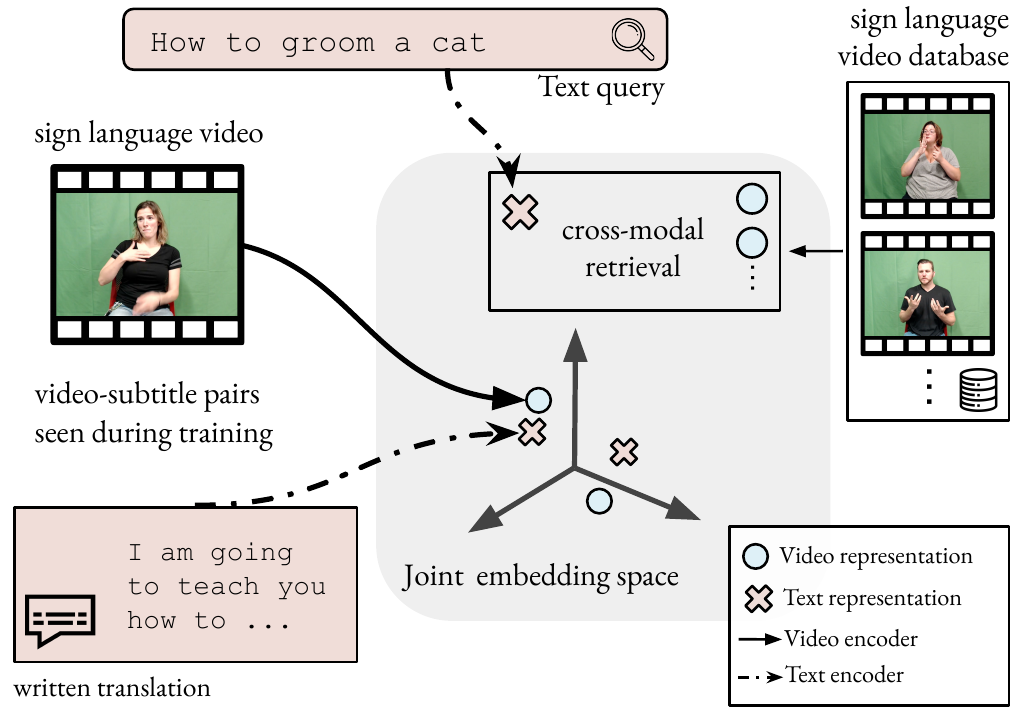}}
    \caption{\textbf{Text-based sign language video retrieval:} In this work we introduce \textit{sign language video retrieval with free-form textual queries},
    the task of searching collections of sign language videos to find the
    best match for a free-form textual query, going beyond individual 
    keyword search.}
    \label{fig:teaser}
\end{figure}

One solution might appear to be to use sign language translation
systems to perform video captioning, analogous to ASR cascading in spoken content retrieval~\cite{Lee2015}.
Unfortunately, while promising translation results have been demonstrated
in constrained domains of discourse (such as weather forecasts)~\cite{camgoz2018neural,camgoz2020sign,li2020tspnet},
it has been widely observed that these systems are unable to achieve functional
performance across the multiple domains of discourse~\cite{bragg2019sign,koller2020quantitative,Varol21}
required for open-vocabulary video indexing (see 
\if\sepappendix1{Appendix~D).}
\else{Appendix~\ref{app:sec:translation}).}
\fi
An alternative solution would be to employ existing methods
for \textit{sign spotting} to perform keyword search.
However, such approaches are fundamentally
brittle---they work best when the user knows exactly
which signs of interest were used in the video.
Moreover, to build an accurate index of such signs using
recent sign spotting techniques~\cite{albanie2020bsl,momeni20_spotting,jiang2021looking}
requires a list of appropriate query candidates,
which to date have often been obtained from subtitles
corresponding to speech transcriptions of the translation,
for example from an ASR engine.
We focus on sign language videos produced by and for signers,
that may not contain any speech track, so producing such speech transcriptions is not an option.

In this work, we address the task of sign language video retrieval with
free-form textual queries by learning a joint embedding space between text and video as illustrated in Fig.~\ref{fig:teaser}.  
Cross-modal embeddings
target only the task necessary to enable search
(i.e.\ ranking a finite pool of sign language videos),
rather than the more involved task of full sign language translation.
As we demonstrate through experiments,
this renders their practical application
even across multiple topics. 
Moreover, cross-modal embeddings enable extremely efficient search
(with the potential to scale up to collections of billions of videos
thanks to mature approximate nearest neighbour algorithms for embedding spaces~\cite{JDH17}).

The task of sign language video retrieval is extremely challenging
for several reasons:
(1) Translation mappings between sign languages and spoken languages
are highly complex~\cite{sutton1999linguistics},
with differing modalities and grammar structures
(ordering is typically not preserved between signed and spoken languages, for example);
(2) In contrast to the datasets used to train text-video retrieval models
(millions of paired examples of videos with corresponding
sentences~\cite{miech2019howto100m,bain2021frozen})
sign language datasets are orders of magnitude smaller in scale;
(3) In addition to a paucity of paired data,
the annotated data available for learning
robust sign embeddings is also extremely scarce
(with sign recognition datasets also considerably smaller
than their counterparts for action recognition~\cite{carreira2019short,ghadiyaram2019large}, for example).

In this work, we propose to study sign language video retrieval on the recently
released \howtosign American Sign Language (ASL) dataset~\cite{duarte2021how2sign}.
To the best of our knowledge,
this dataset represents the largest public source of sign language videos
with aligned captions. 
In order to address the first and second challenges highlighted above,
we construct cross-modal embeddings that leverage pretrained
language models to reduce the burden of data required to learn the mapping between signing sequences and sentences.
To address the third annotation scarcity challenge,
we propose \methodName, a framework for automatic annotation 
that integrates multiple sign spotting methods to
automatically annotate significant fractions of the \howtosign dataset.
By training on the resulting annotations,
we obtain more robust sign embeddings for the downstream retrieval task.

In summary, we make the following contributions:
(1) We introduce the task of \textit{sign language video retrieval with free-form textual queries};
(2) We provide several baselines for this task,
demonstrating the value of cross modal embeddings
and the benefits of incorporating additional retrieval
cues from a sign recognition method 
(whose predictions provide a basis for text-based similarity search)
on the \howtosign and \phoenixT datasets;
(3) We propose the \methodName framework for automatic annotation
and demonstrate its efficacy in producing more robust sign embeddings;
(4) We contribute a new manually annotated test
set for the \howtosign benchmark.
\section{Related Work}
\label{sec:related}

Our work relates primarily to existing research in
\textit{text-video embeddings for video retrieval},
\textit{sign language video retrieval}
and 
\textit{automatic annotation of sign language videos with auxiliary cues},
discussed next.

\noindent \textbf{Text-video embeddings for video retrieval}.
Recently, there has been extensive research interest
in enabling video content search with textual queries via
cross-modal embeddings.
Following the seminal DeViSE model~\cite{frome2013devise}
that demonstrated the strength
of this approach for images and text,
a wide array of text-video embeddings have been explored
~\cite{
Mithun2018LearningJE,yu2018joint,%
miech2018learning,
Liu2019UseWY,wray2019fine,mithun2019weakly,patrick2020support,%
gabeur2020multi,doughty2020action,croitoru2021teachtext,bain2021frozen}.
Differently from these works which target the
retrieval of \textit{describable events},
our work focuses on retrieving
\textit{signing content} that matches a
spoken language query formulated with text.
As noted in Sec.~\ref{sec:intro},
a key challenge that arises from this distinction
is the relative paucity of
training data available to learn a
robust sign video embedding---in this work, we propose
\methodName (introduced in Sec.~\ref{sec:method})
to explicitly address this challenge.

\noindent \textbf{Sign language video retrieval.}
The task of sign language video retrieval
has primarily been investigated 
under the query-by-example search paradigm,
in constrained domains and with small datasets.
In this formulation,
a user query consists of an example of the sign(s)
of interest,
similarly how most keyword-based search engines
deal with text databases.
Two particular variants of this problem have 
received attention for sign language video retrieval:
searching \textit{visual dictionaries of isolated signs},
and searching \textit{continuous sign language datasets},
discussed next.

\noindent \textit{Sign language dictionaries} are video
repositories with recordings of individual signs suitable for learners. 
To search such videos,
Athitsos et al.~\cite{Athitsos2010LargeLP} coupled hand
motion cues with Dynamic Time Warping (DTW)
to enable signer-independent search of an American Sign Language (ASL)
dictionary containing 3k signs and testing with 921 queries.

\noindent For \textit{continuous sign language datasets}, the goal is retrieving
all occurrences of a demonstrated query sign in a target video.
Different techniques have been proposed for this purpose,
including 
hand features with CRFs~\cite{yang2008sign},
hand motion with sequence matching~\cite{Zhang2010UsingRS},
hand and head centroids~\cite{LefebvreAlbaret2010VideoRI},
per-frame geometric features coupled with HMMs~\cite{Zhang2010UsingHT},
and 
non-face skin distribution matching~\cite{viitaniemi2014s}.

As an alternative to querying by example,
a number of works have investigated \textit{sign spotting}
with learned classifiers. 
Ong et al.~\cite{ong2014sign} tackled this problem with HSP-Trees,
a hierarchical data structure built upon Sequential Interval Patterns.
Later work combined human pose estimation with temporal attention
mechanisms to detect (but not localise) the presence of a set 
of glosses among signing sequences~\cite{Tamer2020KeywordSFICASSP}.
This work was later extended to enable search for individual
words~\cite{Tamer2020CrossLingualKS}
and further extended to additionally incorporate hand-shape
features, improving performance~\cite{Tamer2020ImprovingKS}.
More recently, Jiang et al.~\cite{jiang2021looking} showed the
effectiveness of the transformer architecture for the sign spotting task,
achieving promising results on the \bslcorpus~\cite{schembri2013building} and Phoenix2014~\cite{Koller15cslr} datasets.

However, to the best of our knowledge,
no prior sign language retrieval literature
has considered the task that we propose 
in our work, namely \textit{retrieving sign language videos
with free-form textual queries}.

\noindent \textbf{Automatic annotation of sign language with auxiliary cues}.
The abundance of \textit{audio-aligned subtitles}
in broadcast data with sign language interpreters 
has motivated a rich body of work that has sought to
use them as an auxiliary cue to annotate signs.
Cooper and Bowden~\cite{cooper2009learning} propose to use a priori mining to establish
correspondences between subtitles and signs in news broadcasts.
Alternative approaches investigate the use of Multiple Instance
Learning~\cite{buehler2009learning,kelly2010weakly,pfister2013large}.
Other recent contributions leverage words from audio-aligned subtitles
with keyword spotting methods based on mouthing cues~\cite{albanie2020bsl},
dictionaries~\cite{momeni20_spotting}
and attention maps generated by transformers~\cite{Varol21}
to annotate large numbers of signs,
as well as to learn domain
invariant features for improved sign 
recognition through joint training~\cite{Li2020TransferringCK}.

Similarly to these works,
we also aim to automatically annotate sign language videos
by making use of audio-aligned subtitles.
To this end, we make use of prior keyword spotting
methods~\cite{albanie2020bsl,momeni20_spotting}.
However, differently from all the other methods mentioned above
we propose an \textit{iterative} approach,
\methodName,
that alternates between repeated
sign spotting (to obtain more annotations)
and jointly training on the resulting annotations
together with dictionary exemplars (to obtain
better features for spotting).
We note that iterative labelling frameworks have previously seen success in the context of dense sequence re-alignment methods~\cite{koller2017re,koller2019weakly,pu2019iterative} (differently, we target the sparse annotation problem).
We show that our methodology significantly
increases the automatic annotation yield,
and we demonstrate the value of these additional
annotations by using them to learn better representations
for downstream tasks. 
\section{Sign Language Retrieval}
\label{sec:method}

\begin{figure*}[t]
    \centering
            \centering
            \includegraphics[trim={12cm 3cm 0cm 10cm},clip,width=0.95\textwidth]{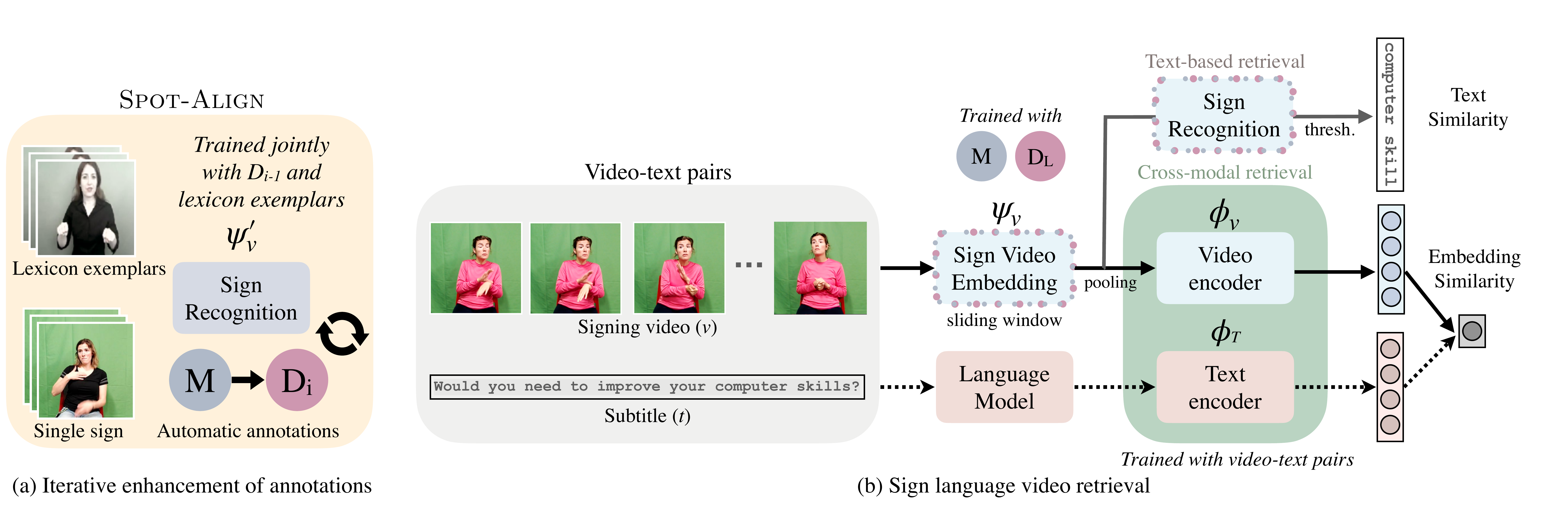}
        \vspace{-0.3cm}
        \caption{\textbf{Method overview:}
        \textit{(a)} We propose \methodName, a framework for
        iteratively increasing annotation yield to obtain
        a good sign video embedding.
        At each iteration $i$,
        the current sign video embedding $\signVideoEmbedding'$
        is trained for classification jointly on the
        \howtosign annotations from iteration $i-1$ 
        and lexicon exemplars from the
        WLASL~\cite{li2020wlasl} and
        MSASL~\cite{joze2018msasl} datasets.
        The resulting improved embedding is then used to
        obtain a new set of sign spottings by re-querying
        \howtosign videos with lexicon exemplars.
        Our final sign video embedding, $\signVideoEmbedding$,
        is obtained by training on the mouthing spottings (M) together with the
        last iteration of dictionary sign spottings, D$_L$
        (without joint training across lexicon exemplars). \td{We provide a detailed sketch of this pipeline in \if\sepappendix1{Appendix~B.1.}
                    \else{Appendix~\ref{app:sec:pipeline}}
                    \fi}
        \textit{(b)} To perform cross modal retrieval, we employ $\signVideoEmbedding$,
        together with a language model, to produce embeddings of videos and text.
        These are passed to a video encoder and text encoder, respectively,
        which are trained to project them into a joint space such that
        they are close if and only if the text matches the video.
        The embedding produced by $\signVideoEmbedding$ is additionally passed
        to a sign recognition model, 
        providing the basis for text-based similarity search.
       \label{fig:framework}
        }
        \vspace{-0.3cm}
\end{figure*}

In this section, we first formulate
the task of 
sign language video retrieval with free-form textual queries
(Sec.~\ref{subsec:method:task}).
Next, we describe the cross-modal (CM) learning formulation considered
in this work (Sec.~\ref{subsec:method:embeddings}),
before introducing our \methodName framework for annotation
enhancement (Sec.~\ref{subsec:method:iterative}).
Finally, we present our text-based retrieval
through our sign recognition (SR) model (Sec.~\ref{subsec:method:recognition}).
Further implementation details are provided in
\if\sepappendix1{Appendix~B.}
\else{Appendix~\ref{app:sec:implementation}.}
\fi

\subsection{Retrieval task formulation}
\label{subsec:method:task}

Let $\videoGallery$ denote a \textit{dataset} of sign language videos of
interest, and let $t$ denote a free-form textual user query.
The objective of the
\textit{sign language video retrieval with textual queries}
task is to find the signing video $v \in \videoGallery$ whose signing content
best matches the query $t$.
We use \textit{text-to-sign-video} (\textToSign) as notation to refer to this task.
Analogously to the symmetric formulations considered in the existing
cross-modal retrieval literature~\cite{dong2016word2visualvec,miech2018learning},
we also consider the reverse \textit{sign-video-to-text} (\signToText) task,
in which a signing video, $v$, is used to query
a collection of text, $\textGallery$.

\subsection{Cross modal retrieval embeddings} 
\label{subsec:method:embeddings}

To address the retrieval task defined above,
we assume access to a parallel corpus of signing videos 
with corresponding written translations.
We aim to learn a pair of encoders, $\signEncoder$ and $\textEncoder$,
which map each signing video $v$ and text $t$ into a common real-valued
embedding space, $\signEncoder(v), \textEncoder(t) \in \mathbb{R}^C$,
such that $\signEncoder(v)$ and $\textEncoder(t)$ are close if and only if
$t$ corresponds to the content of the signing in $v$.
Here $C$ denotes the dimensionality of the common embedding space.

To learn the encoders, we adopt the cross modal ranking learning
objective proposed by Socher~et~al.~\cite{Socher2014GroundedCS}.
Specifically, given paired samples $\{(v_n, t_n)\}_{n=1}^N$,
we optimise a max-margin ranking loss:
\begin{equation}
    \mathcal{L} = \frac{1}{B} \sum_{i=1,i\neq j}^B [\eta_{ij} - \eta_{ii} + m]_+
    + [\eta_{ji} - \eta_{ii} + m]_+
    \label{eqn:margin-loss}
\end{equation}
where $m$ denotes the margin hyperparameter,
$[\cdot]_+$ denotes the hinge function $\text{max}(\cdot, 0)$,
$B$ denotes the size of minibatch sampled during training,
and $\eta_{ij}$ denotes the cosine similarity between signing video $v_i$
and text $t_j$.

Once learned, the embeddings can be applied directly to both the
\textToSign and \signToText tasks.
For the former, inference consists of simply computing the cosine similarity
between the text query $t$ and every indexed signing video $v \in \videoGallery$
to produce a ranking (and vice versa for the \signToText task).

\noindent \textbf{Encoder architectures.}
The sign video encoder, $\signEncoder$ consists first of an initial \textit{sign video embedding},
$\signVideoEmbedding$, which we instantiate as an I3D neural network~\cite{carreira2017quo}
over clips of 16 frames (motivated by its effectiveness for sign recognition tasks~\cite{joze2018msasl,li2020wlasl,albanie2020bsl}).
The output of $\signVideoEmbedding$ is temporally aggregated to a fixed size vector,
and then projected to the $C$-dimensional cross modal embedding space,
$\signEncoder(v) \in \mathbb{R}^C$.

To implement $\textEncoder$, each text sample, $t$, is first embedded through a
language model that has been pretrained on large corpora of written text.
The resulting sequence of word embeddings are then combined
via NetVLAD~\cite{Arandjelovi2018NetVLADCA}
and projected via Gated Embedding Unit
following the formulation of~\cite{miech2018learning} to produce a
fixed-size vector, $\textEncoder(t) \in \mathbb{R}^C$.

In this work, we pay particular attention to the initial
sign video embedding, $\psi_s$, which, as we show through experiments
in Sec.~\ref{sec:experiments}, has a critical influence on performance. 
In Sec.~\ref{sec:experiments}, we also conduct experiments to evaluate
suitable candidates for both the temporal aggregation mechanism on $\signEncoder$,
and the language model employed by $\textEncoder$.

\subsection{Iterative enhancement of video embeddings}
\label{subsec:method:iterative}

As noted above, an effective cross modal embedding for our
task requires a good sign video embedding.
A key challenge in obtaining such an embedding is the relative paucity of annotated sign language data for training.
For example, to the best of our knowledge,
there are no large-scale public datasets of \textit{continuous} signing
with corresponding sign annotations in ASL.

To address this challenge, we propose the \methodName framework
(Fig.~\ref{fig:framework}a)
which we use
to obtain large numbers of \textit{automatic} sign annotations
on the \howtosign dataset.
This dataset provides videos with corresponding written English translations but currently lacks
such annotations.

In summary, we first obtain a collection of candidate sign annotations using
sign spotting techniques proposed in recent works that employ
mouthing cues~\cite{albanie2020bsl} and dictionary
examples~\cite{momeni20_spotting}.
We supplement these sparse annotations:
iteratively increasing the amount of dictionary-based annotations
by retraining our sign video embeddings, and re-querying
dictionary examples. Next, we describe each of these
steps.

\noindent \textbf{Mouthing-based sign spotting~\cite{albanie2020bsl}.}
First, we use the mouthing-based sign spotting
framework of~\cite{albanie2020bsl} to identify sign locations corresponding
to words that appear in the written \howtosign translations.
This approach, which relies on the observation that signing sometimes makes
use of mouthings in addition to head movements and
manual gestures~\cite{sutton1999linguistics},
employs the keyword spotting architecture
of~\cite{stafylakis2018zero} with the improved P2G phoneme-to-grapheme keyword
encoder proposed by Momeni et al.~\cite{Momeni2020SeeingWW}.
We search for keywords from an initial candidate list of 12K words
that result from applying text normalisation~\cite{Flint2017ATN}
to words that appear in \howtosign translations
(to ensure that numbers and dates are converted to their written
form, e.g.\ ``7" becomes ``seven") and filtering to retain only those
words that contain at least four phonemes.
Whenever the keyword spotting model localises a mouthing with a confidence
over 0.5 (out of 1), we record an annotation. With this approach,
we obtain approximately 37K training annotations from a vocabulary of 5K words.
We filter these words to those that appear in the vocabulary of
either WLASL~\cite{li2020wlasl} or MSASL~\cite{joze2018msasl} lexical datasets.
The resulting 9K training annotations cover a vocabulary of 1079 words,
which consists of our initial vocabulary for training a sign
recognition model.

\noindent \textbf{Dictionary-based sign spotting~\cite{momeni20_spotting}.}
Next, we employ an exemplar-based sign spotting method similar to
\cite{momeni20_spotting}. 
This approach considers a handful of video examples
per sign which are used as visual queries to compare against the continuous test video.
The location is recorded as an automatic annotation for the queried sign
at the time where the similarity is maximised.
Such similarity measure between the query and the test videos requires a joint
space. In \cite{momeni20_spotting},
a complex two-stage contrastive training strategy
is formulated.
In this work, we opt for a simpler mechanism in which
we \textit{jointly} train a sign recognition model 
with an I3D backbone, denoted $\signVideoEmbedding'$
on the set of query videos (which are often from an isolated domain
such as lexical dictionaries)
and sign-annotated videos from our search domain
(i.e.\ \howtosign sparse annotations obtained
from the previous step of mouthing-based spotting).
The latent features from this classification model
(which are now approximately aligned between the two domains)
are then used to compute cosine similarities.

Similarly to the mouthing method,
we select candidate query words for each video 
based on the subtitles.
However, when employing dictionary spotting,
we look for both the original and the lemmatised
(removing inflections)
forms of the words, since the sign language lexicons
we employ usually contain a single
version of each word (e.g.\ `run' instead of `running').

As the source of sign exemplars from which we construct queries,
we make use of the training sets of WLASL~\cite{li2020wlasl} and 
MSASL~\cite{joze2018msasl},
two datasets of isolated ASL signing,
with 2K and 1K vocabulary
sizes, respectively.
For joint training, we select samples from their training subsets
that occur in the 1079-sign vocabulary
from our previous mouthing annotations.
However, we use the full training sets for querying,
allowing us to automatically annotate signs outside of the initial 1079 signs.
We record all annotations where the maximum similarity
(over all exemplars per sign) is higher than 0.75 (out of 1),
resulting in 59K training annotations from an expanded vocabulary of 1887 signs.
We initialise the I3D classification from the pretrained
BSL recognition model released by the authors of~\cite{Varol21}.

\noindent\textbf{Iterative enhancement via \methodName.}
From the previous two methods, we obtain an initial set of automatic annotations.
However, the yield of the dictionary-based spotting
method is heavily limited by the \textit{domain gap} between the
videos of \howtosign and the datasets used to obtain the exemplars.
It is therefore natural to ask whether we can improve
the yield from dictionary-based spotting
by achieving a better feature alignment between
the dictionary exemplar
and \howtosign domains.
To this end, we introduce a retrain-and-requery
framework, which we call \methodName, described next.

\begin{figure}
    \centering
    \includegraphics[width=.99\linewidth]{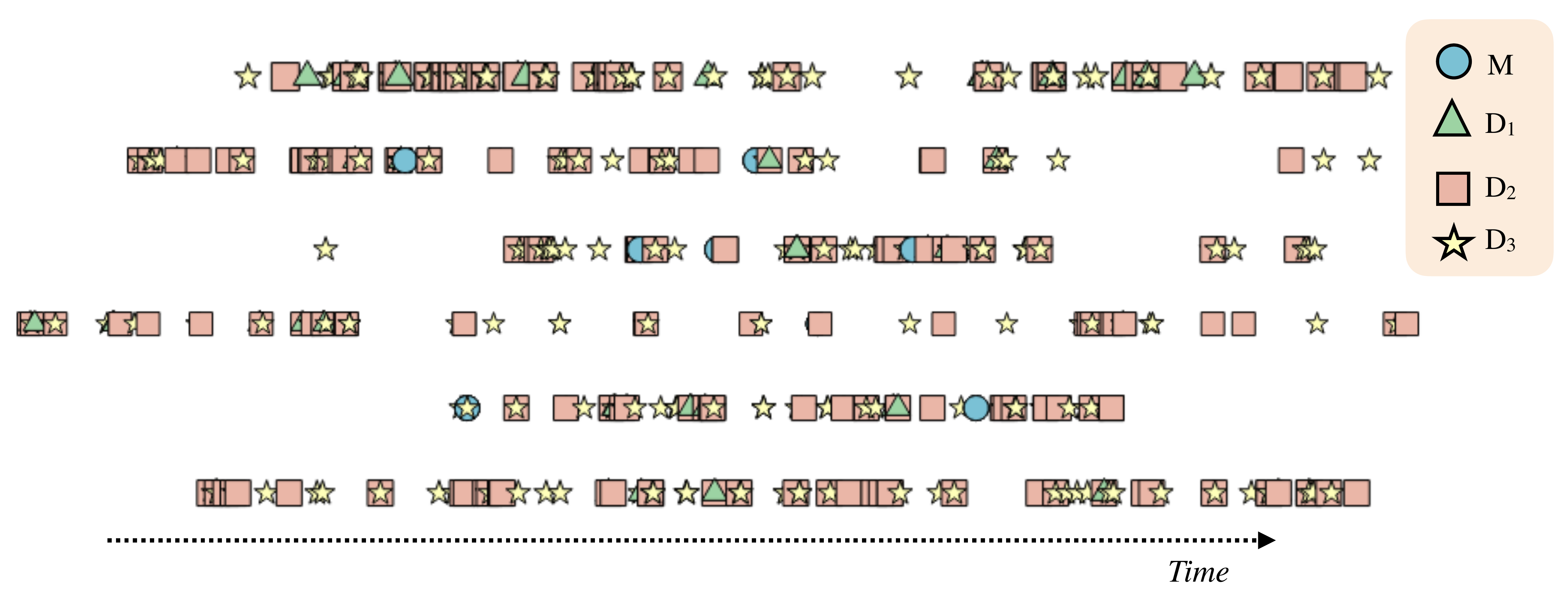}
    \vspace{-0.3cm}
    \caption{\textbf{Iterative enhancement of automatic annotations:} We illustrate sparse annotations generated by different iterations of the \methodName framework on six different video segments (rows) over a fixed-duration interval of 50 seconds each (x-axis).}
    \label{fig:enhancement}
    \vspace{-0.3cm}
\end{figure}

At iteration $i$, we employ the I3D latent features obtained
by joint training between WLASL-MSASL lexicons
and \howtosign automatic annotations provided by iteration $i-1$.
We observe a significant increase in the yield
(e.g.\ 160K annotations in D$_{2}$ vs 59K annotations in D$_{1}$)
despite using the \textit{same exemplars}
and \textit{same subtitles} to construct our queries.
The key difference is then the better aligned
embeddings with which we compare the exemplar and test videos.
In Fig.~\ref{fig:enhancement},
we illustrate the resulting sparse annotations
over a continuous timeline for sample videos
where we observe that
the density of annotations
significantly increases with \methodName iterations.
We denote with D$_{i}$, the set of automatic training annotations
after applying iteration $i$.
An overview of this process is shown in Fig.~\ref{fig:framework}a.

Given the annotations from the final iteration of this process,
we train a new
sign recognition model (trained only on the continuous dataset, i.e.\ \howtosign),
from which we obtain our ultimate video sign embedding $\signVideoEmbedding$
using the (1024-dimensional) latent representation
before the classification layer of 1887 signs.
As shown in Fig.~\ref{fig:framework}b,
this embedding underpins the sign video encoder, $\signEncoder$,
of our cross modal embedding, and is also used to classify individual signs to enable text-based retrieval, described next.

\subsection{Text-based retrieval by sign recognition}
\label{subsec:method:recognition}

The individual sign recognition model used to train the
sign video embedding $\signVideoEmbedding$
can naturally be used to obtain a \textit{sequence} of signs
if applied in a sliding window manner on the long signing videos from $v$.
While the performance of this model is not expected to be high
(due to a lack of temporal modelling stemming from the lack of continuous annotations),
the output list of predicted sign categories gives us a set of candidate words which
can be used to check the overlap with the query text.
This is analogous to
cascading ASR for spoken content retrieval~\cite{Lee2015},
except that sign recognition is significantly
more difficult than speech recognition
(in part, due to a lack of training data~\cite{bragg2019sign}).
Since the order of signs do not necessarily follow word order in the translated text,
we simply compute an Intersection over Union (IoU) to measure
similarity between a query text and the recognised signs.
Before we compute the IoU, we lemmatise both the query words and predicted words.
We constrain the set of recognised signs by removing duplicates and
removing classifications that have probabilities below a certain threshold (0.5 in our experiments).
In Sec.~\ref{sec:experiments}, we show that this text-based retrieval approach,
while performing worse than the cross modal retrieval approach,
is complementary and can significantly boost overall performance.
Implementation details are described in
\if\sepappendix1{Appendix~B.}
\else{Appendix~\ref{app:sec:implementation}.}
\fi

\section{Experiments}
\label{sec:experiments}
We first present the datasets, annotations and evaluation protocols used in our experiments 
(Sec.~\ref{subsec:exp:data}).
Next, we provide retrieval results on \howtosign dataset,
conducting ablation studies to evaluate the influence of different
components of our approach (Sec.~\ref{subsec:exp:how2sign}).
Then, we establish baseline retrieval performances on the \phoenixT
dataset (Sec.~\ref{subsec:exp:phoenix}). 
Finally, we discuss the limitations together with qualitative analysis and societal impact (Sec.~\ref{subsec:exp:limitations}).

\subsection{Data, annotation and evaluation protocols}
\label{subsec:exp:data}
\noindent\textbf{Datasets.} 
In this work, we primarily focus on the recently
released \textbf{\howtosign} 
dataset~\cite{duarte2021how2sign}, a multimodal open-vocabulary and 
subtitled dataset of about 80 hours of continuous sign language videos 
of American Sign Language translation of instructional videos. The 
recorded videos span a wide variety of topics.
We use the videos and their temporally aligned subtitles for training
and evaluating the retrieval model, taking subtitles as textual queries.
There are 31075, 1739 and 2348 
video-subtitle pairs in training, validation and test sets, respectively.
Note that, we remove a small number of videos from the original splits,
where the subtitle alignment is detected
to fall outside the video duration
(more details can be found in
\if\sepappendix1{Appendix~C).}
\else{Appendix~\ref{app:sec:dataset}).}
\fi
We use the validation set to tune parameters (i.e.\ training epoch),
and report all results on the test set.

We also evaluate our sign language retrieval
method to provide baselines on the \textbf{\phoenixT} dataset~\cite{camgoz2020sign},
(although this is not our central focus due to its restricted
domain of discourse).
\phoenixT contains German Sign Language (DGS) videos
depicting weather forecast videos. 
The dataset consists of 7096, 519 and 642 training, validation and test video-text pairs, respectively.
The benchmark is primarily used for sign language translation
where promising results can be obtained due to the restricted
vocabulary size of 3K German words.
Here, we re-purpose it for retrieval, providing baselines using both our cross-modal
embedding approach, and a text-based retrieval by sign language translation \cite{camgoz2020sign}.

\begin{figure}[t]
    \centering
    \includegraphics[height=.49\linewidth]{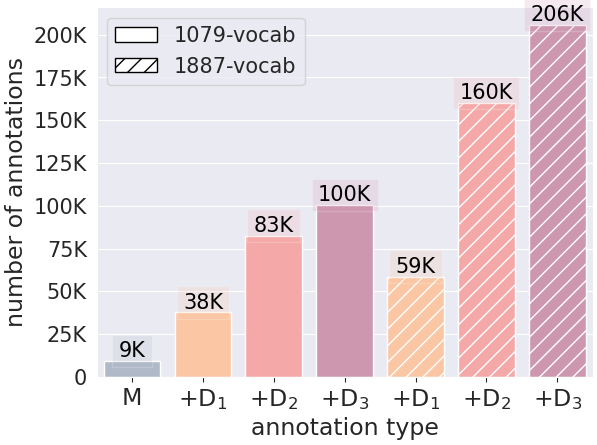}    \raisebox{0.0cm}{\includegraphics[height=.49\linewidth]{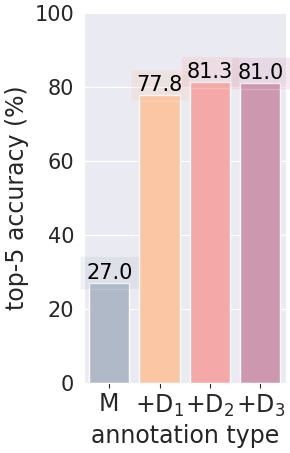}}
    \vspace{-0.3cm}
    \caption{\textbf{Iteratively increasing the sign annotations:}
        Starting from a small set of mouthing annotations,
        we apply sign spotting through dictionaries several times,
        by retraining our I3D backbone on the previous set of automatic
        annotations. The left plot demonstrates the significant increase
        in the number of annotations, for both the restricted (1079)
        and the full (1887) set of categories. The right plot reports
        individual sign recognition (1079-way classification) results on the 
        manually verified test set.
    }
    \label{fig:annotations}
\end{figure}

\begin{table}[t]
    \centering    
    \setlength{\tabcolsep}{2pt}
    \caption{\td{\textbf{Effect of sign video embeddings:} The iterative increase of sign annotations with mouthing- (M) and dictionary-based (D) spotting improves the performance for sign video retrieval tasks with both sign recognition and cross-modal embeddings. The embeddings for the last seven rows are obtained from How2Sign trainings, pretrained
    on BOBSL (second row), which itself was pretrained on Kinetics (first row).}}
    \vspace{-0.3cm}
    \resizebox{0.99\linewidth}{!}{
    \begin{tabular}{lc | cccc | cccc}
        \toprule
         & & \multicolumn{4}{c}{Sign Recognition} & \multicolumn{4}{|c}{Cross-Modal Embeddings} \\
         Sign-Vid-Emb & Vocab & R@1$\uparrow$ & R@5$\uparrow$  & R@10$\uparrow$  & MedR$\downarrow$ & R@1$\uparrow$ & R@5$\uparrow$  & R@10$\uparrow$  & MedR$\downarrow$ \\
         \midrule
         Kinetics~\cite{carreira2017quo} & - & - & - & - & - &
         $1.0_{0.1}$ & $4.4_{0.4}$ & $6.9_{0.6}$ & $296.8_{12.5}$ \\
         BOBSL~\cite{albanie2021bbc} & - & - & - & - & - &
         $17.2_{0.6}$ & $32.5_{0.7}$ & $39.5_{1.3}$ & $30.5_{2.2}$ \\
         \midrule
         M & 1079 & 0.6 & 2.3 & 4.4 & 1174.5 &
         $16.4_{1.2}$ & $31.1_{0.8}$ & $38.2_{0.8}$ & $32.7_{3.1}$ \\
         M+D$_1$ & 1079 & 10.2 & 21.2 & 26.5 & 136.3 &
         $20.6_{1.1}$ & $36.7_{0.6}$ & $43.3_{0.9}$ & $22.0_{2.6}$ \\
         M+D$_2$ & 1079 & 15.6 & 29.0 & 33.9 & 92.0 &
         $21.8_{0.4}$ & $38.0_{0.6}$ & $44.6_{0.8}$ & $18.2_{2.0}$ \\
         M+D$_3$ & 1079 & 16.7 & 29.1 & 33.3 & 95.3 &
         $21.9_{1.2}$ & $38.2_{0.7}$ & $44.8_{0.5}$ & $18.7_{0.6}$\\
         \midrule
         M+D$_1$ & 1887 & 14.1 & 26.1 & 31.4 & 88.0 & 
         $20.4_{0.6}$ & $36.4_{0.3}$ & $43.5_{0.7}$ & $20.0_{1.0}$ \\
         M+D$_2$ & 1887 & 18.3 & 31.3 & 35.8 & 69.8  &
         $23.7_{0.5}$ & $\mathbf{40.8}_{0.1}$ & $\mathbf{47.1}_{0.2}$ & $\mathbf{14.7}_{0.6}$ \\
         M+D$_3$ & 1887 & \textbf{18.4} & \textbf{32.2} & \textbf{36.6} & \textbf{68.0} & 
         $\mathbf{24.5}_{0.2}$ & $40.7_{1.1}$ & $46.7_{0.7}$ & $15.7_{1.5}$\\
         \bottomrule
    \end{tabular}
    }
    \label{tab:features}
\end{table}

\noindent \textbf{Annotations.}
For sign recognition, we train using the automatic sparse annotations
produced by the \methodName framework.
Summary statistics obtained across multiple iterations
of sign spotting are illustrated in Fig.~\ref{fig:annotations} (left),
where we observe a significant increase in yield across consecutive
iterations.
To enable evaluation of sign recognition performance,
we construct a manually verified test set.
This is done by providing annotators proficient in ASL
with sign spotting candidates using the VIA annotation tool~\cite{dutta2019via}.
This results in a recognition test set of 2,212 individual
sign video-category pairs, available at our project page.

\noindent\textbf{Evaluation metrics.}
To evaluate retrieval performance, we follow the
existing retrieval literature~\cite{miech2018learning,Liu2019UseWY,gabeur2020multi}
and report standard metrics R@K (recall at rank K, higher is better) and
MedR (median rank, lower is better).
For sign recognition baselines (for which the time required to
retrain $\signVideoEmbedding$ is more extensive),
we report the results of a single model.
For cross modal embedding ablations (for which the sign video
embedding $\signVideoEmbedding$ is frozen,
and only the text encoder, $\textEncoder$,
and video encoder, $\signEncoder$, are trained),
we report the mean and standard
deviation over three randomly seeded runs.

\begin{figure}[t]
    \centering
    \includegraphics[width=0.48\textwidth]{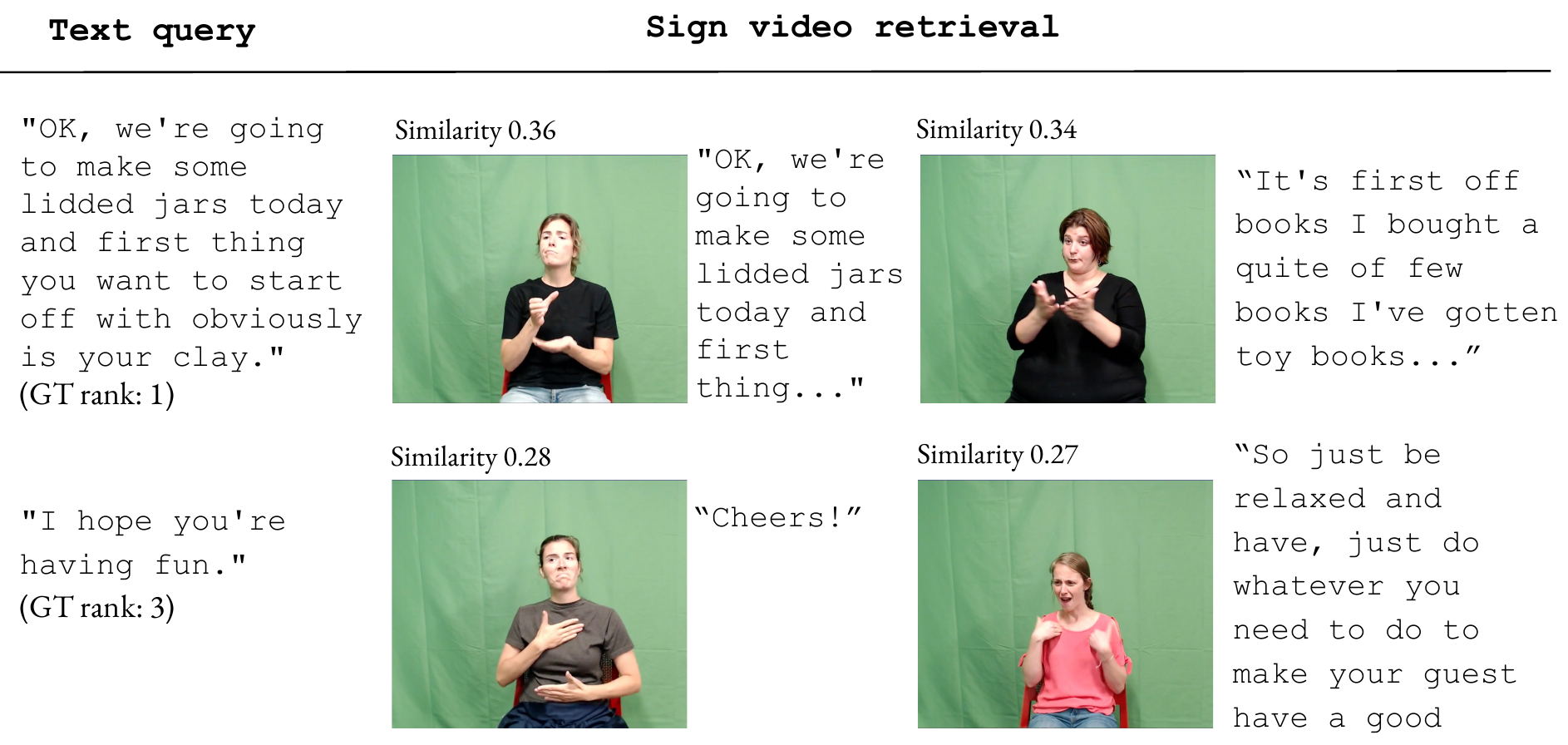}
    \vspace{-0.5cm}
    \caption{\textbf{Qualitative results on text to sign language retrieval:}
    For each query, we show frames from the top two ranked videos as
    well as their corresponding sentences (these are not used during retrieval, and are provided for visualisation purposes). The top row shows a success case.
    The bottom row shows a failure case in which the retrieval model struggles
    with a less detailed query. More examples can be seen in
    \if\sepappendix1{Appendix~A.}
    \else{Appendix~\ref{app:sec:qualitative}.}
    \fi
    }
    \vspace{-0.3cm}
    \label{fig:qualitative}
\end{figure}

\subsection{Retrieval results on How2Sign}
\label{subsec:exp:how2sign}

In this section, we present ablation studies
experimenting with:
(i) different sign video embeddings,
(ii)~video embedding aggregation mechanisms,
and (iii)~text embeddings.
We further study
(iv)~the probability threshold hyperparameter
for text-based retrieval via sign recognition.
We also highlight
(v)~the importance of having a sign language aligned subtitle data
by experimenting with using the original speech-aligned
timings provided by~\cite{duarte2021how2sign}.
Finally, we demonstrate the advantages of
(vi)~combining our cross-modal embedding similarities
with text-based similarities via sign recognition.

\noindent \textbf{(i) Comparison of sign video embeddings.} 
Our main results on sign language retrieval are summarised in 
Tab.~\ref{tab:features}.
Here, we assess the quality of our end-to-end video classification
model to obtain sign video embeddings from the last layer of the I3D model. 
We report both the sign language retrieval from the sign recognition outputs
(using text-to-text matching, as described in Sec.~\ref{subsec:method:recognition}) on the left,
and the learned cross-modal embeddings
(text-to-video matching) on the right.

We first observe that our cross-modal embeddings
(which can potentially capture cues beyond the
limited categories of the sign recognition model)
perform significantly better than their text-based
counterparts.
Next, we compare various choices of backbone
sign video embeddings to evaluate the effectiveness
of our proposed \methodName framework.
As a first baseline, we experiment with using standard
Kinetics~\cite{carreira2017quo} training---we observe
that this produces video embeddings that (as expected)
perform poorly for our task.
We also include as a baseline the model from~\cite{albanie2021bbc}
(pretrained on the BOBSL data)
that was used to initialise our I3D sign video embedding.
\td{The sensitivity of our model to different initialisations is demonstrated in }
\if\sepappendix1{Appendix~D.}
    \else{Appendix~\ref{app:sec:initialization}.}
    \fi
While strongly outperforming Kinetics features,
BOBSL features remain substantially weaker than the
end-to-end ASL sign recognition training \td{on How2Sign} enabled by \methodName.

We observe improvements from each of
our \methodName iterations,
instantiated from mouthing-only (M) annotations,
expanded first in the number of annotations within
the same vocabulary size of 1079,
then expanded in the number of sign categories
with 1887-way classification.
The corresponding statistics for the training size
are illustrated in Fig.~\ref{fig:annotations} (left),
and the sign recognition performance of the corresponding
models on the manually verified test set
can be seen in Fig.~\ref{fig:annotations} (right).
In light of their superior performance,
we use sign video embeddings trained with M+D$_3$ annotations
from the 1887 large-vocabulary for the rest of our experiments on \howtosign.

\noindent\textbf{(ii) Video embedding aggregation.}
Next, we compare the use of different temporal pooling strategies on
the sequence of sign video embeddings for a given sign language video.
While more sophisticated temporal aggregations are possible,
in this work, we opt for a simple and efficient average
pooling mechanism, which has widely been shown to be effective
for text-video retrieval tasks~\cite{miech2018learning,croitoru2021teachtext}.
In Tab.~\ref{tab:feature-aggregation}, we compare average pooling
with maximum pooling over the temporal axis for each feature dimension.
We observe that average pooling performs best.

\begin{table}
    \centering    
    \setlength{\tabcolsep}{6pt}
    \caption{\td{\textbf{Influence of sign video embedding aggregation strategy:}
    We compare temporal pooling strategies on the \howtosign retrieval benchmark.
    Performance metrics are reported as means and standard deviations over
    three randomly seeded runs.}}
    \vspace{-0.3cm}
    \resizebox{0.99\linewidth}{!}{
    \begin{tabular}{l|cccc}
        \toprule
         Aggregation method & R@1$\uparrow$ & R@5$\uparrow$ & R@10$\uparrow$ & MedR$\downarrow$ \\
         \midrule
         Max pooling
         & $23.3_{0.3}$ & $39.7_{0.5}$ & $46.3_{0.6}$ &  $15.3_{0.6}$\\
         Avg. pooling
         & $24.5_{0.2}$ & $40.7_{1.1}$ & $46.7_{0.7}$ & $15.7_{1.5}$ \\

         \bottomrule
    \end{tabular}
    }
    \label{tab:feature-aggregation}
\end{table}




\begin{table}
    \centering    
    \setlength{\tabcolsep}{6pt}
    \caption{\td{\textbf{Influence of the text embedding:}
    We compare a variety of text embeddings on the \howtosign retrieval benchmark.
    Performance metrics are reported as means and standard deviations over
    three randomly seeded runs. }}
    \vspace{-0.3cm}
    \resizebox{0.9\linewidth}{!}{
    \begin{tabular}{l|cccc}
        \toprule
         Text Embedding & R@1$\uparrow$ & R@5$\uparrow$  & R@10$\uparrow$  & MedR$\downarrow$  \\
         \midrule
         GPT~\cite{radford2018improving} 
         & $15.4_{0.4}$ & $30.5_{0.4}$ & $37.6_{0.4}$ & $30.2_{1.3}$ \\
         GPT-2-xl~\cite{Radford2019LanguageMA}
         & $17.0_{0.3}$ & $32.5_{0.4}$ & $39.6_{0.4}$ & $25.7_{1.2}$ \\
         Albert-XL~\cite{Lan2020ALBERTAL}
         & $19.7_{0.3}$ & $36.7_{0.3}$ & $43.8_{0.4}$ & $19.2_{0.8}$ \\
         W2V~\cite{Mikolov2013EfficientEO}
         & $24.2_{0.4}$ & $40.0_{0.4}$ & $46.7_{0.2}$ & $14.8_{0.3}$ \\
         GrOVLE~\cite{burns2019iccv}
         & $24.5_{0.2}$ & $40.7_{1.1}$ & $46.7_{0.7}$ & $15.7_{1.5}$ \\
         \bottomrule
    \end{tabular}
    }
    \vspace{-0.3cm}
    \label{tab:text-embeddings}
\end{table}

\noindent\textbf{(iii) Text embedding.}
We then compare several choices of text embedding for the training of cross modal
embeddings.
We report the results in Tab.~\ref{tab:text-embeddings}.
We observe that word2vec~\cite{Mikolov2013EfficientEO}
and GrOVLE~\cite{burns2019iccv} obtain competitive performance,
outperforming higher capacity alternatives
\cite{radford2018improving,Radford2019LanguageMA,Lan2020ALBERTAL}.
This phenomenon is also observed in \cite{croitoru2021teachtext},
where the authors show that for a number of source text distributions,
simpler word embeddings can outperform their ``heavyweight''
counterparts. We leave the end-to-end fine-tuning of the
language models with our sign language translations to future work,
which can potentially provide further improvements,
and use GrOVLE embeddings for the rest of the experiments.

\noindent\textbf{(iv) Sign recognition probabilities.}
Here, we ablate the text-based retrieval approach
which employs a sign recognition classifier.
Since the sliding window is applied
at each frame densely,
we obtain one sign prediction per frame (which can be very noisy).
Consequently, an important hyperparameter for this method
is the selection of which classification outputs to consider
in our set of predicted words (which will in turn guides the text-based
retrieval). Concretely, the hyperparameter we vary is the confidence threshold
at which predictions are included as text tokens.
We explore several threshold values in Tab.~\ref{tab:threshold} 
and report retrieval performance.
We observe that 0.5 performs best---we adopt this
value for our remaining experiments.

\begin{table}
    \centering    
    \setlength{\tabcolsep}{6pt}
    \caption{\td{\textbf{Thresholding sign recognition probabilities:}
    We investigate the influence of the confidence threshold hyperparameter
    on \howtosign retrieval performance, and observe that 0.5 works best.
    }}
    \vspace{-0.6cm}
    \resizebox{0.7\linewidth}{!}{
    \begin{tabular}{l | cccc}
        \toprule
         Threshold & R@1$\uparrow$ & R@5$\uparrow$ & R@10$\uparrow$ & MedR$\downarrow$ \\
         \midrule
         0.00  & 13.1 & 26.5 & 32.0 & 75.5 \\
         0.10  & 13.4 & 26.4 & 32.4 & 74.0 \\
         0.25 & 17.5 & 30.9 & 35.4 & 56.5 \\
         0.50 & \textbf{18.4} & \textbf{32.2} & \textbf{36.6} & \textbf{68.0}\\
         0.75 & 15.0 & 27.9 & 32.4 & 91.0 \\
         \bottomrule
    \end{tabular}
    }
    \label{tab:threshold}
\end{table}


\noindent\textbf{(v) Effect of alignment.}
We next highlight the importance of having aligned video-sentence pairs,
motivating our selection of \howtosign
(which contains the largest public source of aligned video-sentence pairs)
as a test-bed for our study of large-vocabulary retrieval.
We retrain cross modal embeddings on speech-aligned training data,
and evaluate both the recognition and cross modal embedding models
on the speech-aligned test data of \howtosign to compare with
signing-aligned version.
The results are reported in Tab.~\ref{tab:alignment},
where we observe that speech-aligned subtitles significantly
damage retrieval performance.

\begin{table}
    \centering    
    \setlength{\tabcolsep}{6pt}
    \caption{\textbf{Effect of subtitle alignment:}
    We report retrieval performance on \howtosign for models
    trained and evaluated on subtitles aligned to speech and
    to signing.
    We observe a significant drop in performance when using speech-aligned
    subtitles.
    }
    \vspace{-0.3cm}
    \resizebox{0.99\linewidth}{!}{
    \begin{tabular}{l | cccc | cccc}
        \toprule
         & \multicolumn{4}{c}{Sign Recognition} & \multicolumn{4}{c}{Cross-Modal Embeddings} \\
         Alignment & R@1$\uparrow$ & R@5$\uparrow$  & R@10$\uparrow$  & MedR$\downarrow$ & R@1$\uparrow$ & R@5$\uparrow$  & R@10$\uparrow$  & MedR$\downarrow$ \\
         \midrule
         Speech & 9.5 & 16.1 & 19.0 & 418.0 &
         ${5.9}_{0.6}$ & ${13.6}_{0.6}$ & ${18.0}_{0.2}$ & ${483.5}_{17.9}$ \\
         Signing & \textbf{18.4} & \textbf{32.2} & \textbf{36.6} & \textbf{68.0} & 
         $\mathbf{24.5}_{0.2}$ & $\mathbf{40.7}_{1.1}$ & $\mathbf{46.7}_{0.7}$ & $\mathbf{15.7}_{1.5}$\\
         \bottomrule
    \end{tabular}
    }
    \label{tab:alignment}
\end{table}



\noindent\textbf{(vi) Combining several cues.}
Finally, in Tab.~\ref{tab:ensembling}, we combine our two types of models
based on sign recognition (SR) and cross-modal embeddings (CM).
We perform \textit{late fusion} (averaging the similarities,
with equal weights) computed with individual models.
We conclude that sign recognition provides complementary cues
to our cross-modal embedding training, significantly boosting
the final performance. Tab.~\ref{tab:ensembling} presents both 
\textToSign and \signToText performances establishing a new benchmark
on the task of retrieval for the recent \howtosign dataset. Some qualitative examples of videos retrieved by our system are provided in Fig.~\ref{fig:qualitative}.

\begin{table}
    \centering    
    \setlength{\tabcolsep}{6pt}
    \caption{\td{\textbf{Combination of models:}
    We report our final benchmark performance on \howtosign
    for the retrieval models based on sign recognition
    (SR) and cross modal (CM) embeddings.
    We observe that the two approaches are highly complementary.
    }}
    \vspace{-0.3cm}
    \resizebox{0.99\linewidth}{!}{
    \begin{tabular}{l | cccc | cccc}
        \toprule
         & \multicolumn{4}{c}{\textToSign} & \multicolumn{4}{|c}{\signToText} \\
         Models & R@1$\uparrow$ & R@5$\uparrow$  & R@10$\uparrow$  & MedR$\downarrow$& R@1$\uparrow$ & R@5$\uparrow$  & R@10$\uparrow$  & MedR$\downarrow$\\
         \midrule
         SR & 18.4 & 32.2 & 36.5 & 68.0 & 11.5 & 27.9 & 33.3 & 66.0\\
         CM & 24.7 & 39.6 & 46.0 & 17.0 & 17.9 & 40.8 & 46.6 & 15.0 \\
         SR + CM & \textbf{32.8} & \textbf{47.7} & \textbf{52.9} & \textbf{7.0} & \textbf{23.3} & \textbf{48.5} & \textbf{53.7} & \textbf{7.0} \\
         \bottomrule
    \end{tabular}
    \vspace{-0.3cm}
    }
    \label{tab:ensembling}
\end{table}





\subsection{Retrieval results on \phoenixT}
\label{subsec:exp:phoenix}

In addition to the \howtosign ASL dataset that formed the primary basis
of our study, we also provide retrieval baselines on the \phoenixT
dataset~\cite{Koller15cslr,camgoz2018neural}.
For cross modal embedding training, 
we employ a text embedding model
trained on German language corpora,
GPT-2~\cite{Radford2019LanguageMA} 
released by Chan~et~al.~\cite{Chan2020GermansNL}.
For text-based retrieval, here we incorporate a state-of-the-art sign language
translation model~\cite{camgoz2020sign}, with which we compute an IoU similarity
measure. Note that sign language translation performance is high on this
dataset due to its restricted domain of discourse, which is the reason
why we opt for a translation-based approach instead of the recognition-based
retrieval as in Sec.~\ref{subsec:method:recognition}.
\td{In both experiments we use the video features provided by~\cite{camgoz2020sign,koller2019weakly}.}
The results are reported in Tab.~\ref{tab:phoenx14T}.
We observe that our cross-modal embeddings strongly outperform
the translation-based retrieval. Their combination
performs best (as in Tab.~\ref{tab:ensembling}).

\begin{table}
    \centering    
    \setlength{\tabcolsep}{6pt}
    \caption{\textbf{Retrieval performance on the \phoenixT dataset:}
    We report baseline performances for cross modal embeddings, as
    well as text-based retrieval by sign language translation
    on the 642 sign-sentence pairs of the test set.
    }
    \vspace{-0.3cm}
    \resizebox{0.99\linewidth}{!}{
    \begin{tabular}{l|cccc|cccc}
        \toprule
         & \multicolumn{4}{c}{\textToSign} & \multicolumn{4}{c}{\signToText} \\
         Text Embedding & R@1$\uparrow$ & R@5$\uparrow$ & R@10$\uparrow$ & MedR$\downarrow$ & R@1$\uparrow$ & R@5$\uparrow$ & R@10$\uparrow$ & MedR$\downarrow$ \\
         \midrule
        Translation~\cite{camgoz2020sign} & 30.2 & 53.1 & 63.4 & 4.5 & 28.8 & 52.0 & 60.8 & 56.1 \\
        Cross-modal & 48.6 & 76.5 & 84.6 & 2.0 & 50.3 & 78.4 & 84.4 & \textbf{1.0} \\
        \midrule
        Combination & \textbf{55.8} & \textbf{79.6} & \textbf{87.2} & \textbf{1.0} & \textbf{53.1} & \textbf{79.4} & \textbf{86.1} & \textbf{1.0} \\
         \bottomrule
    \end{tabular}
    }
    \vspace{-0.3cm}
    \label{tab:phoenx14T}
\end{table}






\subsection{Limitations and societal impact}
\label{subsec:exp:limitations}

One limitation of \methodName method is \edit{that it is} not able
to discover new signs
outside of the vocabulary of queried lexicons.
Qualitatively, we observe failure cases of our cross-modal retrieval
model (illustrated in Fig.~\ref{fig:qualitative}),
when using more generic queries that lack precise detail.
\td{
It is also worth noting that all datasets used in our experiments are interpreted sign language rather than conversational 
(e.g.\ conversations between native signers -- see~\cite{bragg2019sign} for a broader discussion on how this can limit models trained on such data).}

\noindent \textbf{Societal Impact.} The ability to efficiently search sign language videos
has a number of useful applications for content creators
and researchers in the deaf community.
However, by providing this technical capability, it also
potentially brings risk of surveillance of signers, since large volumes of signing content can be searched automatically.
\section{Conclusion}
\label{sec:conclusion}

In this work, we introduced the task of
sign language video retrieval with free-form textual queries.
We provided baselines for this task on the \howtosign
and \phoenixT datasets.
We also proposed the \methodName framework to obtain automatic
annotations, and demonstrated their value in producing effective sign video embeddings for retrieval.

{\footnotesize
\noindent \textbf{Acknowledgements.}
The authors would like to thank M.~Fischetti, C.~Marsh and M.~Dippold for their work on data annotation and also A. Dutta, A. Thandavan, J.~Pujal, L.~Ventura, E.~Vincent, L.~Tarres, P.~Cabot, C.~Punti and Y.~Kalantidis for their help and valuable feedback.
This work was supported by the project PID2020-117142GB-I00, funded by MCIN/ AEI /10.13039/501100011033,
ANR project CorVis ANR-21-CE23-0003-01, and gifts from Google and Adobe.
AD received support from la Caixa Foundation (ID 100010434), fellowship code LCF/BQ/IN18/11660029.
SA thanks Z. Novak and N. Novak for enabling his contribution.
\par
}

{\small
\bibliographystyle{ieee_fullname}
\bibliography{references}
}

\clearpage
\bigskip
{\noindent \large \bf {APPENDIX}}\\
\renewcommand{\thefigure}{A.\arabic{figure}} 
\setcounter{figure}{0} 
\renewcommand{\thetable}{A.\arabic{table}}
\setcounter{table}{0} 

\appendix

This appendix provides additional qualitative analyses (Sec.~\ref{app:sec:qualitative}),
implementation details (Sec.~\ref{app:sec:implementation}), dataset details (Sec.~\ref{app:sec:dataset}), \td{additional experiments demonstrating the sensitivity of our model to different initialisations (Sec.~\ref{app:sec:initialization}),}
and an experiment demonstrating challenges of using text-based retrieval via sign language translation (Sec.~\ref{app:sec:translation}).

\section{Qualitative Analysis}
\label{app:sec:qualitative}

\noindent\textbf{Supplemental webpage.} We qualitatively illustrate, 
in our project page, 
 (\url{https://imatge-upc.github.io/sl_retrieval/app-qualitative/index.html}),
the retrieval results using the best model on the How2Sign
dataset (SR+CM combination from 
\if\sepappendix1{Tab.~6).}
\else{Tab.~\ref{tab:ensembling}).}
\fi
For each query, we show the top three ranked videos as well as their corresponding
topic category (see \cite{duarte2021how2sign}
for more details of video topic categories),
signer ID and sentences (note that these are not used during retrieval, and are provided for visualisation  purposes).

The top ten rows of the webpage show cases in which our
model is able to correctly retrieve the video corresponding
to the textual query.
The middle five rows of the webpage show cases where the correct video is not
retrieved successfully.
For these failures, we nevertheless observe that the retrieval model
makes reasonable mistakes
(for instance, in the majority of cases,
at least one of the top three ranked videos
share the same topic category of the GT video).
In the bottom five rows, we show examples of
failure cases of our model.

\begin{figure*}
    \centering
    \includegraphics[width=\textwidth]{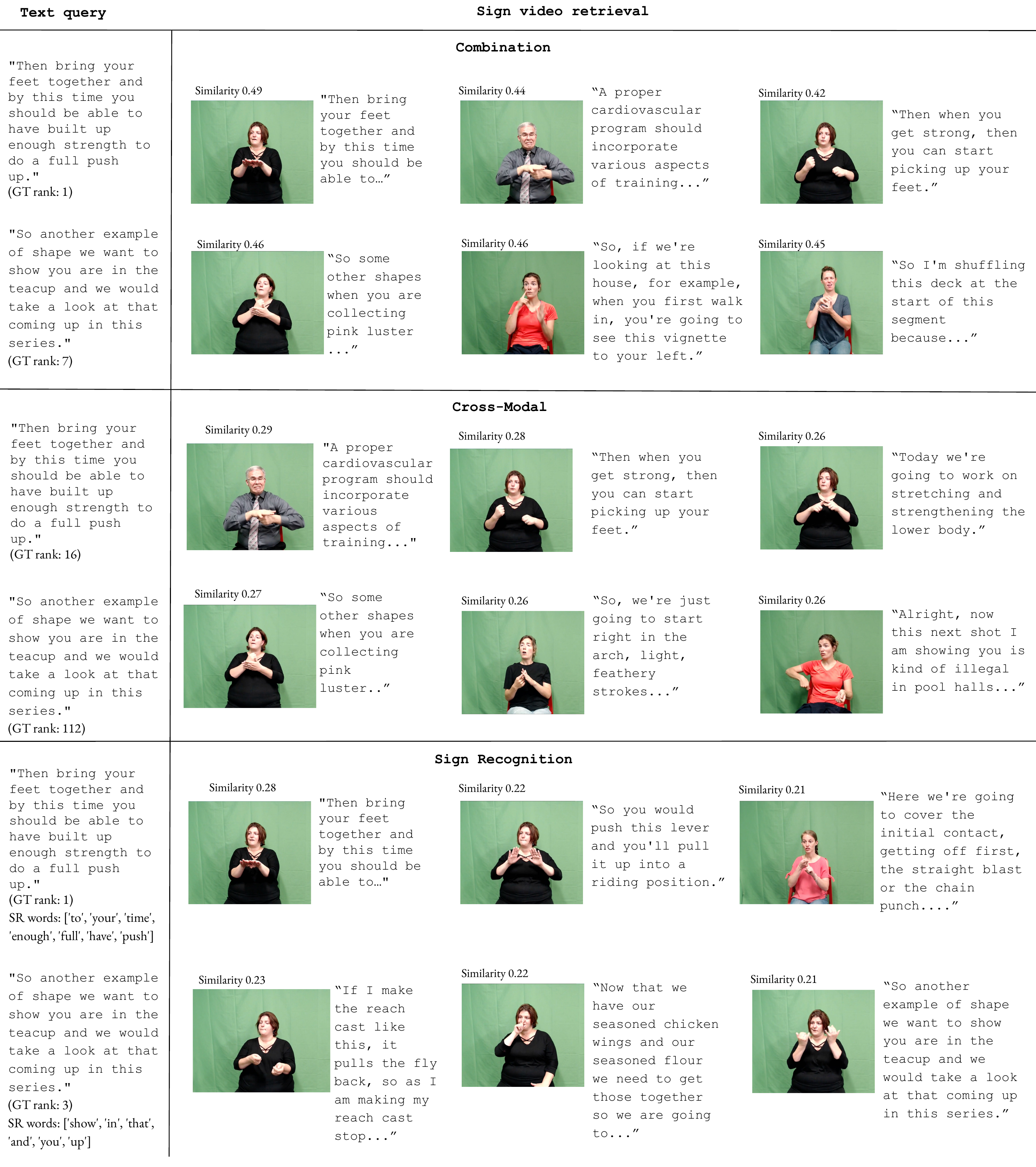}
    \caption{\textbf{Qualitative results:} We show two samples where text-based retrieval using sign recognition (SR) helps retrieval when combined with cros modal embeddings (CM). Top, middle and bottom rows show the retrieval
    results for the same query using the average of the similarities from SR and CM (Combination), Cros Modal and Sign Recognition models, respectively. 
    }
    \vspace{0.3cm}
    \label{app:fig:qualitative_results_breakdown}
\end{figure*}

\noindent\textbf{Combination of cross modal and sign recognition.}
We noticed that 
our strongest retrieval model combines similarities
from the cross modal embeddings and the sign recognition model
(SR+CM combination from 
\if\sepappendix1{Tab.~6}
\else{Tab.~\ref{tab:ensembling}}
\fi
).
In Fig.~\ref{app:fig:qualitative_results_breakdown}, we illustrate
two example queries for which the use of the sign recognition model
substantially improves the performance of the cross modal embeddings.

\section{Implementation Details}
\label{app:sec:implementation}

In this section, 
\td{we provide a detailed sketch of the \methodName pipeline (Sec.~\ref{app:sec:pipeline}), as well as the} 
additional implementation details
for the sign video embedding (Sec.~\ref{sec:sign-video-embedding:impl}),
text embedding (Sec.~\ref{sec:text-embedding:impl})
and
cross modal retrieval training (Sec.~\ref{sec:cross-modal:impl}).

\begin{figure}
    \centering
    \resizebox{.99\linewidth}{!}{
    \includegraphics[]{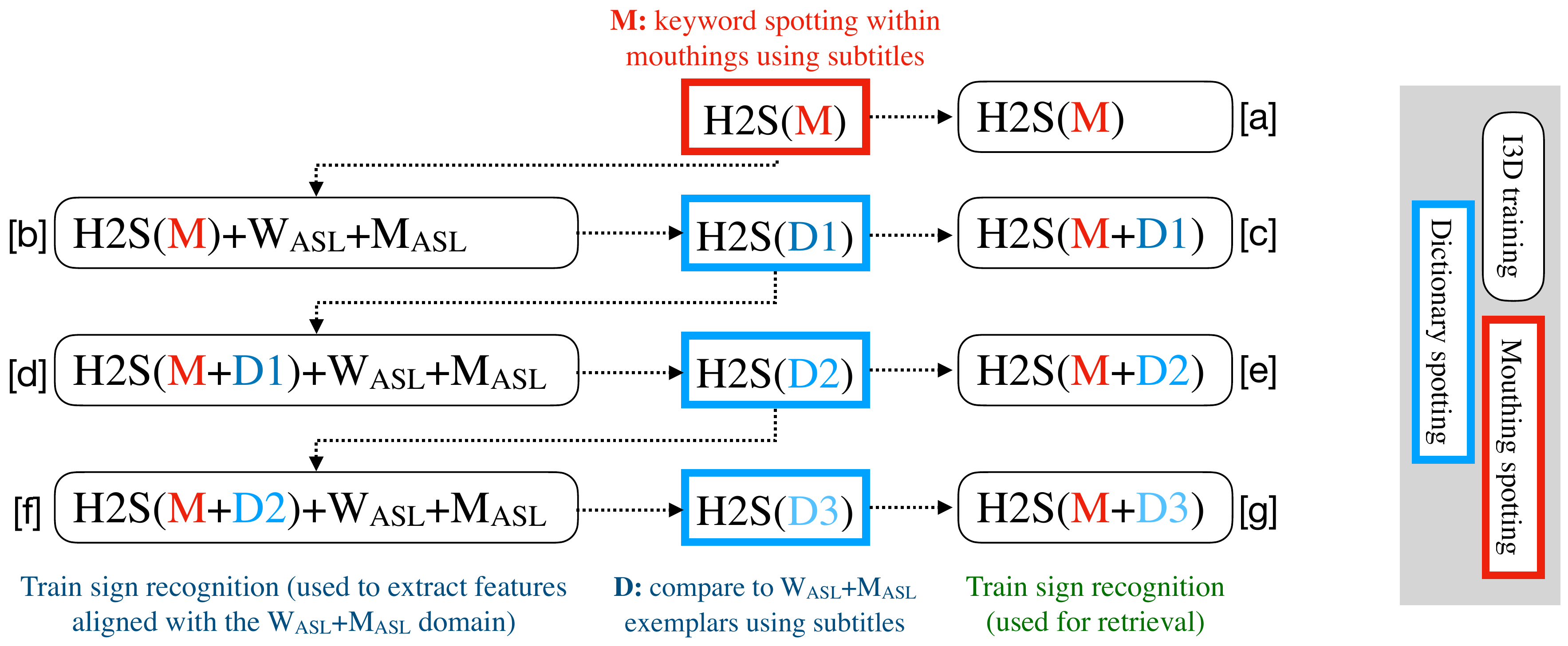}}
    \caption{\td{\textbf{Pipeline sketch of \methodName iterations:} [a] We use a Mouthing-based sign spotting to obtain an initial set of automatic sign-level annotations on the How2Sign (H2S) dataset which we call here H2S(M). [b] Using the automatic annotations obtained, we jointly train on the continuous signing examples from H2S(M) and the dictionary-style signing videos from WLASL and MSASL, in order to obtain a feature space aligned between
    the two domains. A Dictionary-based sign spotting approach
    is then used to obtain a new set of sign spottings (D1) by re-querying How2Sign videos with lexicon exemplars.
    The process is then iterated with the new spottings, as described in the main paper.}}
    \label{app:fig:framework}
     \vspace{-0.4cm}
\end{figure}

\td{
\subsection{SPOT-ALIGN iterations}
\label{app:sec:pipeline}
In Fig.~\ref{app:fig:framework} we provide a detailed sketch of the \methodName framework and how we obtain our Mouthing and Dictionary-based annotations for the How2Sign dataset.
On the left ([b], [d], [f]), we show the iterations for the joint training
between How2Sign, WLASL and MSASL, which is used for Dictionary-based
sign spotting. On the right ([a], [c], [e], [g]), the training is only performed on the How2Sign dataset,
which provides sign video embeddings for retrieval.
}
\subsection{Sign recognition and sign video embedding} 
\label{sec:sign-video-embedding:impl}

\noindent\textbf{Sign recognition training.}
As explained in 
\if\sepappendix1{Sec.~3.4}
\else{Sec.~\ref{subsec:method:recognition}}
\fi
of the main paper, we train a
sign recognition model, a 3D convolutional neural network
instantiated with an I3D~\cite{carreira2017quo} architecture
\td{pretrained on BOBSL~\cite{albanie2021bbc}}. 
We finetune this model on the How2Sign dataset using our automatic sign
spotting annotations. In the final setting with mouthing (M)
and dictionary (D$_3$) spottings from a vocabulary of 1887 signs,
we have 206K training video clips, each corresponding to a single
sign. Since the spottings represent a point in time, rather than a segment
with beginning-end times, we determine a fixed window for each
video clip. 
For mouthing annotations, this window is defined as 15 frames before the annotation time and 4 frames after ($[-15, 4]$). For dictionary annotations, the window is similarly set to $[-3, 22]$. During training, we randomly sample
16 consecutive frames from this window, such that the RGB video input to the network
becomes of dimension $16 \times 3 \times 224 \times 224$. We apply
a similar spatial cropping randomly from $256 \times 256$ resolution.
We further employ augmentations such as colour jittering, resizing and horizontal flipping.

We perform a total of 25 epochs on the training data,
starting with a learning rate of 1e-2, reduced by a factor of
10 at epoch 20. We optimise using SGD with momentum (with a value of 0.9)
and a minibatch of size 4.

At test time, for recognition, we apply a sliding window averaging in time, and center cropping in space. 
At test time, for text-based retrieval,
we obtain the predicted class per 16-frame sliding window (with a stride of 1 frame),
and record the corresponding word out of the 1887-vocabulary if the probability
is above the 0.5 threshold. The resulting set of words are merged
in case of repetitions, and are compared against the queried
text to obtain an intersection over union (IOU) score, used
as the similarity.

\noindent\textbf{Sign video embedding.}
As noted in
\if\sepappendix1{Sec.~3.2}
\else{Sec.~\ref{subsec:method:embeddings}}
\fi
of the main paper, we employ
the I3D recognition model (described above)
to instantiate our sign video embedding.
More specifically, we use the outputs corresponding to the
spatio-temporally pooled vector
before the last (classification) layer. This produces
a 1024-dimensional real-valued vector for each 16 consecutive RGB frames.
We extract these features densely with a stride 1 from How2Sign
sign language sentences to obtain the sequence of sign video embeddings.


\subsection{Text embedding}
\label{sec:text-embedding:impl}
We consider several text embeddings in this work.
When conducting experiments on the \howtosign dataset,
we explore the following English language embeddings: 

\vspace{4pt}

\noindent \textbf{GPT}~\cite{radford2018improving} is a
768-dimensional embedding that uses a Transformer
decoder which is trained on the
BookCorpus~\cite{Zhu2015AligningBA} dataset.

\noindent \textbf{GPT-2-xl}~\cite{Radford2019LanguageMA}
is a 1600-dimensional embedding
(employing 1558M parameters, also in a
Transformer architecture~\cite{vaswani2017attention})
that is trained on the WebText corpus (containing millions
of pages of web text).

\noindent \textbf{Albert-XL}~\cite{Lan2020ALBERTAL}
is a 2048-dimensional embedding that builds on
BERT~\cite{Devlin2019BERTPO} to increase its efficiency.
It is trained with a loss that models inter-sentence
coherence on the BookCorpus~\cite{Zhu2015AligningBA}
and Wikipedia~\cite{Devlin2019BERTPO} datasets.

\noindent \textbf{W2V}~\cite{Mikolov2013EfficientEO}
is a 300-dimensional word embedding,
trained on the Google News corpus (we use the
\texttt{GoogleNews-vectors-negative300.bin.gz}
model from \url{https://code.google.com/archive/p/word2vec/}).

\noindent \textbf{GroVLE}~\cite{burns2019iccv}.
This is a 300-dimensional embedding that aims to
be vision-centric:
it is adapted from Word2Vec~\cite{Mikolov2013EfficientEO}

\vspace{4pt}

For experiments on the \phoenixT dataset, we use a German
language model:

\vspace{4pt}

\noindent \textbf{German GPT-2}~\cite{Chan2020GermansNL}
(based on the original GPT-2 architecture
of ~\cite{radford2018improving}) is a 768-dimensional
embedding. The model is trained on
the OSCAR~\cite{OrtizSuarez2019AsynchronousPF}
corpus, together with a blend of smaller German language
data.
We use the parameters made available at
\url{https://huggingface.co/dbmdz/german-gpt2}.

\subsection{Cross modal retrieval} 
\label{sec:cross-modal:impl}

The dimensionality of the shared embedding space
(denoted by the variable $C$ in 
\if\sepappendix1{Sec.~3.2}
\else{Sec.~\ref{subsec:method:embeddings}}
\fi
)
used in this work is 512.
The margin hyperparameter, $m$, introduced in 
\if\sepappendix1{Eqn.~1,}
\else{Eqn.~\ref{eqn:margin-loss},}
\fi
is set to 0.2,
following~\cite{miech2018learning}.
All cross modal embeddings are trained for 40 epochs
using the RAdam optimiser~\cite{Liu2019OnTV} with a learning
rate of 0.001, a weight decay of $1E-5$ and a batch size of 128.
For each experiment, the epoch achieving the highest geometric
mean of R@1, R@5 and R@10 on the validation set was used to select
the final model for test set evaluation.
The NetVLAD~\cite{Arandjelovi2018NetVLADCA} layer employed in the
text encoder uses 20 clusters.
Sign video embeddings
(which form the input to the video encoder, $\signEncoder$
described in 
\if\sepappendix1{Sec.~3.2}
\else{Sec.~\ref{subsec:method:embeddings}}
\fi
)
are extracted densely (i.e.\ with a temporal stride 1).

\section{Dataset Details}
\label{app:sec:dataset}

To construct training, validation and test retrieval partitions
from the \howtosign dataset,
we select video segments with their corresponding manually
aligned subtitles (released by the authors of~\cite{duarte2021how2sign}).
This provides an initial pool of
31,164 training,
1,740 validation
and 2,356 test videos
with corresponding translations.
After initial inspection, we found that while most annotations
were produced to a high quality, a small
number of the manually aligned subtitles were invalid
(i.e.\ exhibited no temporal overlap with the video).
We excluded these invalid subtitles from our retrieval
benchmark, producing final splits of:
31,075 training,
1,739 validation
and 2,348 test videos.

\begin{figure}
    \centering
    \includegraphics[width=0.49\linewidth]{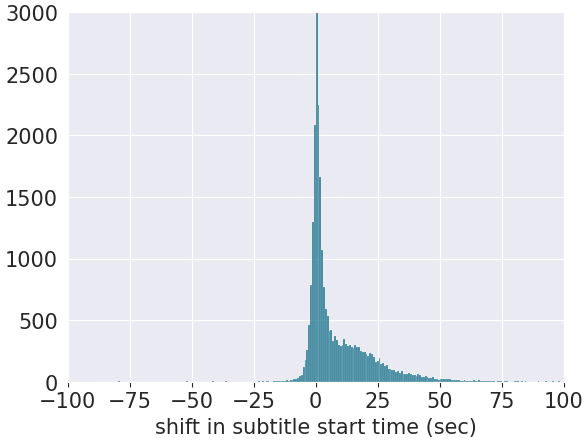}
    \includegraphics[width=0.49\linewidth]{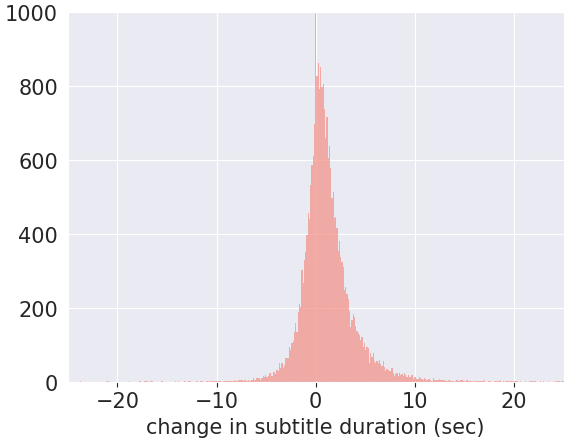}
    \caption{We plot two histograms to illustrate the difference between the sign-aligned
    and speech-aligned subtitles. On the left, we show the distribution of $t_{beg}^{sign} - t_{beg}^{speech}$,
    i.e.\ the shift between the beginning subtitle times between the sign- and speech-aligned versions.
    On the right, we similarly plot the distribution of changes in durations. The peaks
    at the zero-bin are at 7,000 and 5,000 for left and right plots, respectively, which are
    truncated for better visibility.
    }
    \label{app:fig:alignment}
\end{figure}

In Fig.~\ref{app:fig:alignment},
we visualise the difference between the timings of the original subtitles
versus the manually aligned subtitles. We note that the signing
is on average behind the speech, constituting a misalignment when using
the original subtitle timings.
This misalignment explains the performance drop we demonstrated in
\if\sepappendix1{Tab.~5}
\else{Tab.~\ref{tab:alignment}}
\fi
of the main paper
when experimenting with the original subtitles instead of the manually aligned ones.

\begin{table}[t]
    \centering    
    \caption{\textbf{Sensitivity to initialisation:} \td{We investigate the effects of different initialisation for our sign video embedding. We experiment with random and WLASL initialisation. D$_{1,\textcolor{red}{\text{BSL1K}},\textcolor{blue}{ft\text{(H$_{M}$WM)}}}$
    means obtaining D$_{1}$ by pretraining the [b] model (see Fig.~\ref{app:fig:framework})
    on BSL-1K~\cite{Varol21} and finetuning jointly on H2S mouthing annotations and WLASL/MSASL
    exemplars.}}
    \label{tab:supmat_init}
    \vspace{-0.3cm}
    \setlength{\tabcolsep}{2pt}
    \resizebox{0.99\linewidth}{!}{
    \begin{tabular}{ll | lc | cccc | cccc}
        \toprule
         & & \#tr.& Acc. & \multicolumn{4}{c}{Sign Recognition} & \multicolumn{4}{|c}{Cross-modal retrieval} \\
         Sign-Vid-Emb & Init [a][c] & ann. & top-5 & R@1$\uparrow$ & R@5$\uparrow$  & R@10$\uparrow$  & MedR$\downarrow$ & R@1$\uparrow$ & R@5$\uparrow$  & R@10$\uparrow$  & MedR$\downarrow$ \\
         \midrule
         M                  & BOBSL & 9K & 27.0
                                    & 0.6 & 2.3 & 4.4 & 1174.5
                                    & $16.4_{1.2}$ & $31.1_{0.8}$ & $38.2_{0.8}$ & $32.7_{3.1}$ \\
         M+D$_{1,\textcolor{red}{\text{BSL1K}},\textcolor{blue}{ft\text{(H$_{M}$WM)}}}$
                                    & BOBSL & 38K & 77.8
                                    & 10.2 & 21.2 & 26.5 & 136.3
                                    & $20.6_{1.1}$ & $36.7_{0.6}$ & $43.3_{0.9}$ & $22.0_{2.6}$ \\
         \midrule
         M                  & BSL-1K & 9K & 25.1 & 0.6 & 2.4 & 4.4 & 1174.5 & $18.0_{0.7}$ & $32.4_{0.6}$ & $39.3_{0.7}$ & $27.8_{1.6}$ \\ 
         M+D$_{1,\textcolor{red}{\text{BSL1K}},\textcolor{blue}{ft\text{(H$_{M}$WM)}}}$
                                    & BSL-1K & 38K & 77.7 & 10.4 & 22.6 & 27.6 & 131.5 & $20.8_{0.8}$ & $36.9_{0.9}$ & $43.5_{0.8}$ & $20.5_{0.5}$ \\
         \midrule
         M                          & WLASL   & 9K & 23.5
                                    & 1.1 & 2.9 & 4.2 & 1175.5
                                    & $11.3_{0.5}$ & $23.0_{0.5}$ & $29.5_{0.6}$ & $67.3_{7.2}$ \\
         M+D$_{1,\textcolor{red}{\text{WLASL}},\textcolor{blue}{ft\text{(H$_{M}$WM)}}}$   
                                    & WLASL & 60K & 72.7
                                    & 9.3 & 19.9 & 24.5 & 208.8
                                    & $17.1_{0.6}$  & $31.5_{0.6}$ & $38.3_{0.4}$ & $32.8_{2.5}$ \\
        \midrule
         M                & random & 9K & 6.5
                                    & 0.0 & 0.0 & 0.0 & 1174.5 
                                    & $0.6_{0.1}$ & $2.0_{0.2}$ & $3.3_{0.5}$ & $530.7_{16.3}$ \\
         M+D$_{1,\textcolor{red}{\text{random}},\textcolor{blue}{ft\text{(H$_{M}$WM)}}}$
                                    & random & 136K & 28.0
                                    & 0.0 & 0.5 & 0.8 & 1175.0
                                    & $2.4_{0.2}$  & $7.2_{0.2}$ & $10.2_{0.0}$ & $221.3_{6.4}$ \\
         \bottomrule
    \end{tabular}
    \vspace{-0.4cm}
    }
\end{table}

\td{
\section{Sensitivity to Initialisation}
\label{app:sec:initialization}
We provide in Tab.~\ref{tab:supmat_init} comparisons for training with
annotations from Mouthing (M) and the first iteration of Dictionary (D1)
spottings
([a] and [c] in Fig.~\ref{app:fig:framework})
from four different initialisations: I3D weights pretrained on BOBSL~\cite{albanie2021bbc},
BSL-1K~\cite{Varol21},
WLASL~\cite{li2020wlasl},
or randomly initialised.
Note that all BOBSL, BSL-1K and WLASL models are also initialised from Kinetics.
Here, we rerun the Dictionary-based sign spotting to obtain
different sets of D1 annotations
by initialising from WLASL-pretrained and random weights
(instead of BSL-1K model from~\cite{Varol21} in the rest
of the experiments).
While random initialisation significantly hurts performance, the WLASL-pretrained model performs slightly worse than~\cite{albanie2021bbc}, demonstrating that our method
can work provided a reasonable initialisation.
Assuming access to WLASL is realistic
since we use it in step [b].}

The performances of BOBSL versus BSL-1K pretraining are
similar in Tab.~\ref{tab:supmat_init}. Our preliminary
results also suggest that similar trends from
\if\sepappendix1{Tab.~1}
\else{Tab.~\ref{tab:features}}
\fi
hold
when pretraining all rows on BSL-1K instead of BOBSL.
We therefore report all our models ([a, c, e, g]) with BOBSL pretraining since this dataset~\cite{albanie2021bbc} has recently become available (unlike the BSL-1K source data~\cite{albanie2020bsl}, which is not public).
However, we clarify that
the Dictionary spottings were obtained with models ([b,~d,~f])
pretrained on BSL-1K.

Furthermore, 
we investigate whether the domain alignment between
WLASL+MSASL exemplars and How2Sign is beneficial
by comparing
M+D$_{1,\textcolor{red}{\text{BSL1K}},\textcolor{blue}{ft\text{(H$_{M}$WM)}}}$
and
M+D$_{1,\textcolor{red}{\text{BSL1K}}}$. The latter
consists of 112K spottings (as opposed to 38K); however,
the top-5 recognition accuracy drops to 60.0\% (from 77.7\%)
suggesting the poor quality of feature alignment between
the two domains in the absence of joint finetuning.

\section{Text-based Retrieval Attempt through Sign Language Translation}
\label{app:sec:translation}

As mentioned in 
\if\sepappendix1{Sec.~1}
\else{Sec.~\ref{sec:intro}}
\fi
of the main paper, a text-based retrieval solution
using sign language translation on videos is not a viable option due to
unsatisfactory video-to-text translation performance of current state-of-the-art
models~\cite{camgoz2020sign} on open-vocabulary domains. Here, we provide a brief
justification by training the encoder-decoder Transformer model of \cite{camgoz2020sign}
on the How2Sign dataset using the same sign video embeddings as in our cros modal
retrieval setting, i.e.\ the densely extracted I3D features. We keep all the hyperparameters
identical to the publicly available setting of \cite{camgoz2020sign} and obtain
poor BLEU scores of 1.74 and 17.08 for BLEU-4 and BLEU-1, respectively. We provide in our the project page
(\url{https://imatge-upc.github.io/sl_retrieval/app-translation/index.html})
the ground truth (left)
and the predicted (right) sentences on the validation
set and observe that the predictions tend to be generic sentences that do not correspond
to the input sign language video, with the exception that sometimes the model
predicts one word right out of the entire sentence.
\end{document}